\documentclass[runningheads,orivec,a4paper]{llncs}

\usepackage{amssymb}
\usepackage{graphicx}
\usepackage{mathtools}
\usepackage{textcomp}
\usepackage{amsmath}
\usepackage{amsopn}
\usepackage{tabularx, booktabs}
\usepackage{array}
\newcolumntype{Y}{>{\centering\arraybackslash}X}
\newcolumntype{C}[1]{>{\centering\let\newline\\\arraybackslash\hspace{0pt}}m{#1}}
\DeclareMathOperator{\Softmax}{Softmax}

\begin{document}

\mainmatter  

\title{Generalised Wasserstein Dice Score for Imbalanced Multi-class Segmentation\\
      using Holistic Convolutional Networks}

\titlerunning{Generalised Wasserstein Dice Score}

\author{Lucas Fidon\inst{1}%
	\and Wenqi Li\inst{1}\and Luis C. Garcia-Peraza-Herrera\inst{1}\and\\ 
	Jinendra Ekanayake\inst{2,3}\and Neil
        Kitchen\inst{2}\and\\S\'ebastien Ourselin\inst{1,3} \and Tom Vercauteren\inst{1,3}}

%
\authorrunning{Lucas Fidon et al.}

\institute{TIG, CMIC, University College London, London, UK\\
	\and
	NHNN, University College London Hospitals, London UK\\
	\and
	Wellcome / EPSRC Centre for Interventional and Surgical Sciences, \\UCL, London, UK}

\maketitle

\begin{abstract}
The Dice score is widely used for binary segmentation due to its robustness to class imbalance.
Soft generalisations of the Dice score allow it to be used as a loss function for training convolutional neural networks (CNN).
%
%
%
Although CNNs trained using mean-class Dice score achieve state-of-the-art
results on multi-class segmentation, this loss function
does neither take advantage of inter-class relationships nor
multi-scale information.
We argue that an improved loss function should balance
misclassifications to favour predictions that are
semantically meaningful.  
%
%
This paper investigates these issues in the context of multi-class brain tumour segmentation.
Our contribution is threefold. 1) We propose a semantically-informed
generalisation of the 
Dice score for multi-class segmentation based on the Wasserstein
distance on the probabilistic label space.
2) We propose a holistic CNN that embeds spatial information at multiple scales with deep supervision.
%
3) We show that the joint use of holistic CNNs and generalised
Wasserstein Dice score achieves segmentations that are more semantically meaningful for brain tumour segmentation. 
\end{abstract}

\section{Introduction}

\begin{figure}[t!]
	\centering
	\includegraphics[width=0.78\linewidth]{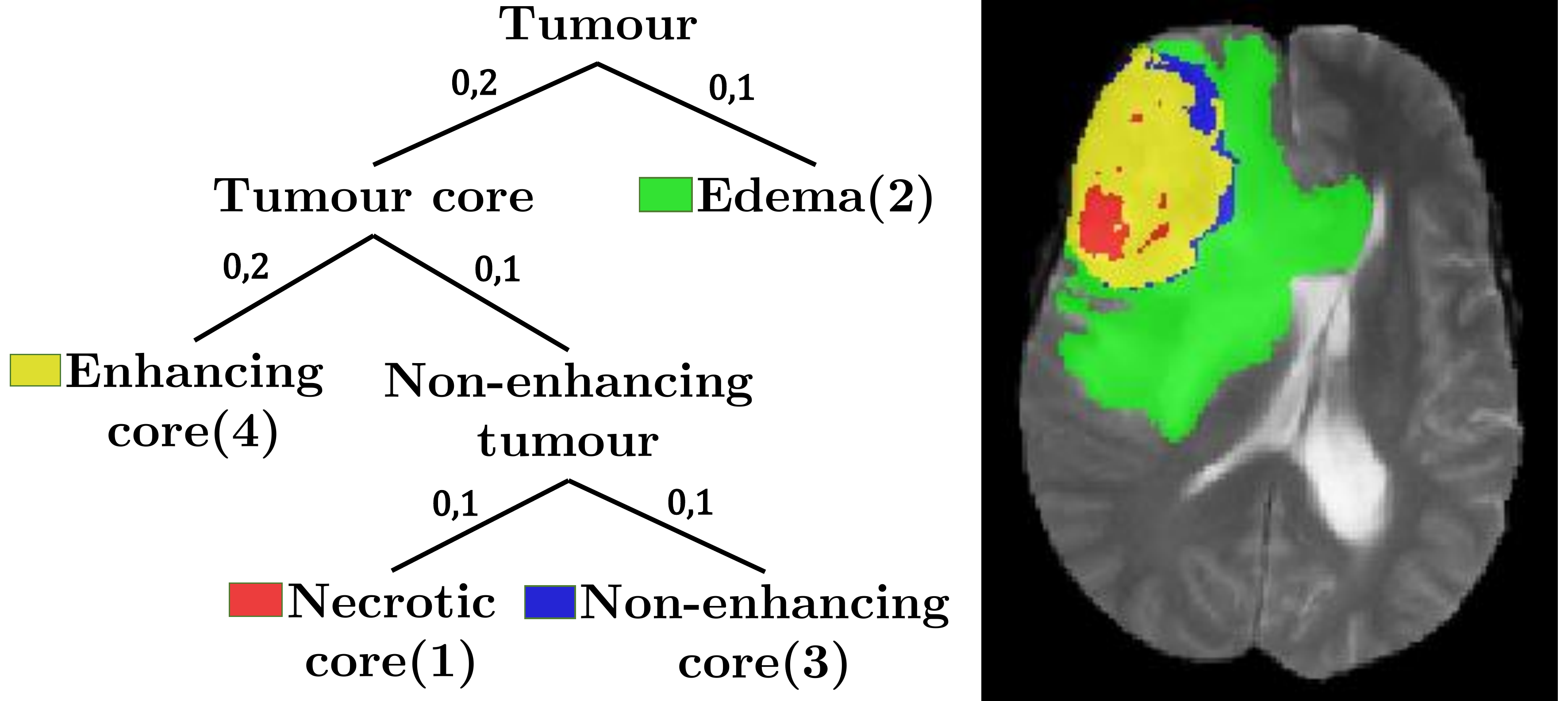}
	\caption{Left: tree on BraTS label space. Edge weights have been manually selected to reflect the distance between labels. Right: illustration on a T2 scan from BraTS'15~\cite{brats}.}
	\label{fig:tree}
\end{figure}

Automatic brain tumour segmentation is an active research area.
Learning-based methods using convolutional neural networks (CNNs) have recently emerged as the state of the art~\cite{Havaei2017,deep_medic}.
%
%
One of the challenges is the severe class imbalance.
%
Two complementary ways have traditionally been used when training CNNs to tackle
imbalance: 1) using a sampling strategy that imposes
constraints on the selection of image patches; and 2) using pixel-wise weighting
to balance the contribution of each class in the objective function. 
%
%
%
For CNN-based segmentation, samples should ideally be entire
subject volumes to support the use of fully convolutional network and
maximise the computational efficiency of
convolution operations within GPUs.  
As a result, weighted loss functions appear more promising to improve
CNN-based automatic brain tumour
segmentation. 
%
%
Using soft generalisations of the Dice score (a popular
overlap measure for binary segmentation) directly as a loss function has
recently been proposed~\cite{v_net,Sudre2017}. By introducing global spatial
information into the loss function, the Dice loss has been shown to be
more robust to class imbalance.  
%
%
However at least two sources of information are not fully utilised in this formulation: 1) the
structure of the label space; and 2) the spatial information  across scales.
%
Considering the class imbalance and the hierarchical label structure illustrated in Fig.\ref{fig:tree}, both of them are likely to play an important role for multi-class brain
tumour segmentation. 

In this paper, we propose two complementary contributions that leverage
prior knowledge about brain tumour structure. 
First, we exploit the Wasserstein distance~\cite{wasserstein_loss,fast_emd}, which can
naturally embed semantic relationships between classes for the
comparison of label probability vectors, to generalise the Dice score
for multi-class segmentation.
%
%
Second, we propose a new holistic CNN architecture inspired
by~\cite{toolnet,hed} that embeds spatial information at different
scales and introduces deep supervision during the CNN training.
%
%
%
We show that the combination of the proposed generalised Wasserstein Dice score and our Holistic CNN achieves better generalisation
compared to both mean soft Dice score training and classic CNN architectures
for multi-class brain tumour segmentation. 

\section{A Wasserstein approach for multi-class soft Dice score}

\subsection{Dice score for crisp binary segmentation}
The Dice score is a widely used overlap measure for pairwise comparison of binary segmentations $S$ and $G$. It can be expressed both in terms of set operations or statistical measures as:
\begin{equation}\label{dice}
D = \frac{2|S \cap G|}{|S| + |G|} = \frac{2\Theta_{TP}}{2\Theta_{TP} + \Theta_{FP} + \Theta_{FN}} = \frac{2\Theta_{TP}}{2\Theta_{TP} + \Theta_{AE}}
\end{equation}
with $\Theta_{TP}$ the number of true positives, $\Theta_{FP}$/$\Theta_{FN}$ the number of false positives/false negatives, and $\Theta_{AE}=\Theta_{FP}+\Theta_{FN}$ the number of all errors.

\subsection{Dice score for soft binary segmentation}
Extensions to soft binary
segmentations~\cite{Anbeek2005,chang2009performance} rely on the
concept of disagreement for pairs of probabilistic classifications.
The classes $S_i$ and $G_i$ of each voxel $i \in \mathbf{X}$ can be defined
as random variables on the label space $\mathbf{L}=\{0,1\}$ and the
probabilistic segmentations can be represented as label probability
maps: $p=\{p^i:=P(S_i=1)\}_{i \in \mathbf{X}}$ and
$g=\{g^i:=P(G_i=1)\}_{i \in \mathbf{X}}$. 
We denote $P({\mathbf{L}})$ the set of label
probability vectors.
We can now generalise $\Theta_{TP}$ and $\Theta_{AE}$ to soft
segmentations:
\begin{equation}\label{soft_dsc}
\Theta_{AE} = \sum_{i \in \mathbf{X}} |p^i - g^i|, \quad
\Theta_{TP} = \sum_{i \in \mathbf{X}}g^i(1 - |p^i - g^i|)
\end{equation}
In the common case of a crisp segmentation $g$ (i.e. $\forall i \in \mathbf{X}, g^i \in \{0,1\}$), the associated soft Dice score can be expressed as:
\begin{equation}
D(p,g) = \frac{2\sum_i g^ip^i}{\sum_i (g^i + p^i)}
\end{equation}
A second variant has been used in~\cite{v_net}, with a quadratic term in the denominator.

\subsection{Previous work on multi-class Dice score}
The easiest way to derive a unique criterion from the soft binary Dice score for multi-class segmentation is to consider the mean Dice score:
\begin{equation}\label{mean_dice}
 D_{mean}(p,g) = \frac{1}{|\mathbf{L}|}\sum_{l \in \mathbf{L}}\frac{2\sum_i g^i_lp^i_l}{\sum_i (g^i_l + p^i_l)}
\end{equation}
where $\{g^i_l\}_{i \in \mathbf{X},\, l \in \mathbf{L}}$, $\{p^i_l\}_{i \in \mathbf{X},\, l \in \mathbf{L}}$ are the set label probability vectors for all voxels for the ground truth and the prediction.

A generalised soft multi-class Dice score has also been
proposed in~\cite{gdsc,Sudre2017} by generalising the set theory definition of
the Dice score \eqref{dice}: 
\begin{equation}\label{gdsc}
D_{FM}(p,g) = \frac{2\sum_{l}\alpha_{l}\sum_{i}\min(p^i_l,g^i_l)}{\sum_{l}\alpha_{l}\sum_{i}(p^i_l + g^i_l)}
\end{equation}
where $\{\alpha_l\}_{l \in \mathbf{L}}$ allows to weight the
contribution of each class. However, those definitions are still based
only on pairwise comparisons of probabilities associated with the same
label and don't take into account inter-class relationships.

\subsection{Wasserstein distance between label probability vectors}
The Wasserstein distance (also sometimes called the \emph{Earth
	Mover's Distance}) represents the minimal cost to transform a
probability vector $p$ into another one $q$ when for all $l,l' \in
\mathbf{L}$, the cost to move a unit from $l$ to $l'$ is defined as
the distance $M_{l,l'}$ between $l$ and $l'$. 
This is a way to map a distance matrix $M$
(often referred to as the \emph{ground distance matrix})
on
$\mathbf{L}$, into a distance on $P({\mathbf{L}})$ that leverages
prior knowledges about $\mathbf{L}$. 
In the case of a finite set $\mathbf{L}$, for $p, q \in
P({\mathbf{L}})$, the Wasserstein distance between $p$ and $q$ derived
from $M$ can be defined as the solution of a linear programming
problem~\cite{fast_emd}:
\begin{equation}\label{wasserstein}
\begin{split}
&W^M(p,q) = \min_{T_{l,l'}} \sum_{l,l' \in \mathbf{L}}
T_{l,l'}M_{l,l'}, \\
\textrm{subject to }	&\forall l \in  \mathbf{L}, \, \sum_{l' \in
	\mathbf{L}}T_{l,l'} = p_l,\, \textrm{and} \quad
\forall l' \in  \mathbf{L}, \, \sum_{l \in \mathbf{L}}T_{l,l'} =
q_{l'}.
\end{split}
\end{equation}
where $T=(T_{l,l'})_{l,l' \in \mathbf{L}}$ is a joint probability distribution for $(p,q)$ with marginal distributions $p$ and $q$.
%
A value $\hat{T}$ that minimises (\ref{wasserstein}) is called an
\emph{optimal transport} between $p$ and $q$ for the distance
matrix $M$.

\subsection{Soft multi-class Wasserstein Dice score}
The Wasserstein distance $W^M$ in \eqref{wasserstein} yields a natural way to 
compare two label probability vectors in a semantically meaningful
manner by supplying a distance matrix $M$ on $\mathbf{L}$.
%
Hence we propose using it
to generalise the
measure of disagreement between a pair of label probability vectors
and provide the following generalisations:
\begin{align}\label{wass_AE}
\Theta_{AE} &= \sum_{i \in \mathbf{X}} W^M(p^i, g^i) \\
\Theta_{TP}^l &= \sum_{i \in \mathbf{X}}g^i_l(W^M(l, b) - W^M(p^i, g^i)),
                \quad \forall l \in \mathbf{L}\setminus\{b\}
\end{align}
where $W^M(l, b)$ is shorthand for $M_{l,b}$ and $M$ is
chosen such that the background class $b$ is always the furthest away
from the other classes. 
To generalise $\Theta_{TP}$, we propose to
weight the contribution of the classes similarly to
\eqref{gdsc}: 
\begin{equation}\label{wass_TP}
\Theta_{TP} = \sum_{l \in \mathbf{L}} \alpha_{l} \,\Theta_{TP}^l
\end{equation}
We chose
$\alpha_{l}=W^M(l, b)$ to make sure that background voxels do not contribute to $\Theta_{TP}$.
\noindent The Wasserstein Dice score with respect to $M$ can then be defined as:
\begin{equation}\label{generalised dice}
D^M(p, g) = \frac{ 2\sum_{l} W^M(l, b)\sum_{i} g^i_l(W^M(l, b) - W^M(p^i, g^i))}{2\sum_{l}[ W^M(l, b)\sum_{i} g^i_l(W^M(l, b) - W^M(p^i, g^i)) ] + \sum_{i} W^M(p^i, g^i)}
\end{equation}
In the binary case, setting  
$M= \begin{bsmallmatrix}
0       & 1  \\
1       & 0  \\
\end{bsmallmatrix}$ leads to $W^M(p^i,g^i)=|p^i-g^i|$ and reduces the proposed
Wasserstein Dice score to the soft binary Dice score
\eqref{soft_dsc}. 

\subsection{Wasserstein Dice loss with crisp ground truth}
Previous work on Wasserstein distance-based loss functions for deep
learning have been limited because of the computational
burden~\cite{fast_emd}. However, in the case of a crisp
ground-truth $\{g^i\}_i$, and for any prediction $\{p^i\}_i$, a
closed-form solution exists for \eqref{wasserstein}.
An optimal transport is $\forall l,l' \in \mathbf{L},
\hat{T}_{l,l'}=p^i_lg^i_{l'}$ and the Wasserstein distance becomes: 
\begin{equation}\label{wass_closed-form}
W^M(p^i, g^i) = \sum_{l,l' \in \mathbf{L}} M_{l,l'}p^i_lg^i_{l'}
\end{equation}
We define the Wasserstein Dice loss derived from $M$ as $\mathcal{L}_{D^M}:=1-D^M$.
\section{Holistic convolutional networks for multi-scale fusion}
\begin{figure}[b!]
	\centering
	\includegraphics[width=\linewidth]{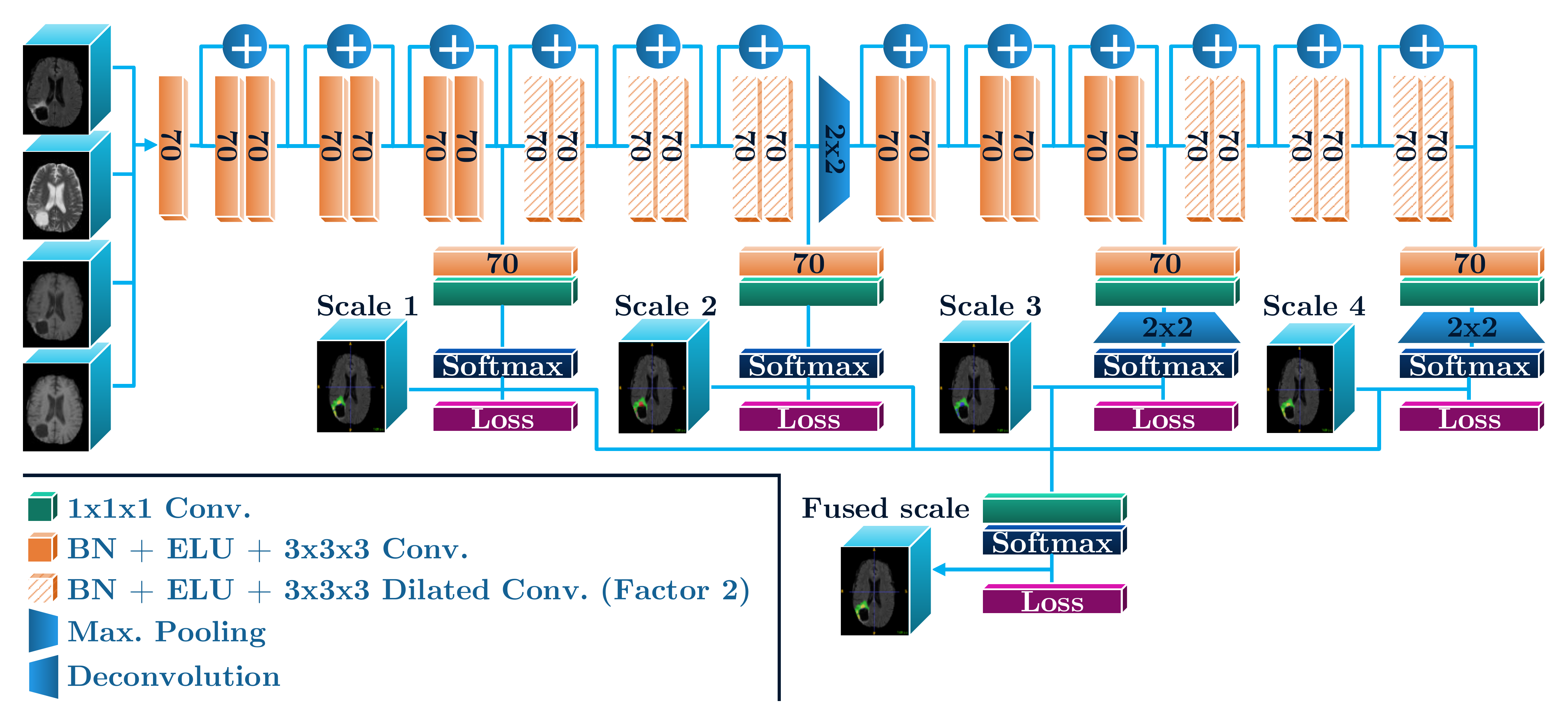}
	\caption{Proposed holistically-nested CNN for multi-class labelling of brain tumours.}
	\label{fig:hcnn_architecture}
\end{figure}

We now describe a holistically-nested convolutional neural network
(HCNN) for imbalanced multi-class brain tumour segmentation inspired by
the holistically-nested edge detection (HED) introduced
in~\cite{hed}. HCNN has been used successfully for some imbalanced
learning tasks such as edge detection in natural images~\cite{hed} and
surgical tool segmentation~\cite{toolnet}.
The HCNN features multi-scale prediction and intermediate
supervision. It can produce a unified output using a fusion layer
while implicitly embedding spatial information in the loss. 
We further improve on the ability of HCNNs to deal with imbalanced
datasets by leveraging the proposed generalised Wasserstein Dice loss. 
To keep up with state-of-the-art CNNs, we also employ ELU as
activation function~\cite{elu} and use residual
connections~\cite{resnet}. Residual blocks include a pair of $3^3$
convolutional filters and Batch Normalisation~\cite{wideresnet}. The
proposed architecture is illustrated in
Fig.\ref{fig:hcnn_architecture}.

\subsection{Multi-scale prediction to leverage spatial consistency in the loss}
%
%
%
%
%
As the receptive field 
increases across successive layers, predictions computed
at different layers embed spatial information at different scales. 
Especially for imbalanced multi-class segmentation,
different scales can contain complementary information. 
%
In this paper, to increase the receptive field and avoid redundancy between
successive scale predictions, max pooling and
dilated convolutions (with a factor of $2$ similar to~\cite{highresnet})
have been used. 
As predictions are computed
regularly at intermediate scales along the network (Fig.\ref{fig:hcnn_architecture}), we chose to increase the number of features before the first prediction is made. For simplicity reasons, we then selected
the same value for all hidden layers (fixed to $70$ given memory
constraints). 

\subsection{Multi-scale fusion and deep supervision for multi-class segmentation}
While classic CNNs provide only one output, HCNNs
provide outputs $\hat{y}^{s}$ at $S$ different layers of the network,
and combine them to provide a final output $\hat{y}^{fuse}$: 
\[
(\hat{y}_{l}^{fuse})_{l \in \mathbf{L}} = \Softmax\Big(\big(\sum_{s=1}^{S} w_{l,s} \hat{y}_{l}^{s}\big)_{l \in \mathbf{L}}\Big).
\]
As different scales can be of different importance for different
classes we learn class-specific fusion weights $w_{l,s}$. 
This transformation can also be represented by a convolution layer
with kernels of size $1^3$ where the multi-scale
predictions are fused in separated branches for each class, as
illustrated in Fig.~\ref{fig:hcnn_architecture} similarly to the
scalable layers introduced in~\cite{scalenet}. 
%
In addition to applying the loss function $\mathcal{L}$ to the fused
prediction, $\mathcal{L}$ is also applied to each scale-specific
prediction thereby providing deep supervision (coefficients $\bar{\lambda}$ and $\lambda_{s}$ are set to 1/$(S+1)$ for simplicity):
\[
\mathcal{L}_{Total}((\hat{y}^{s})_{s=1}^{S}, \hat{y}^{fuse}, y) = \bar{\lambda}\mathcal{L}(\hat{y}^{fuse}, y) + \sum_{s=1}^{S} \lambda_{s}\mathcal{L}(\hat{y}^{s}, y)
\]
%


\section{Implementation details}

\subsection{Brain tumour segmentation}
We evaluate our HCNN model and Wasserstein Dice loss functions on the
task of brain tumour segmentation using BraTS'15 training set that
provides multimodal images (T1, T1c, T2 and Flair) for 220 high-grade
gliomas subjects and 54 low-grade gliomas subjects.  We divide it
randomly into $80\%$ for training, $10\%$ for validation and $10\%$
for testing so that the proportion of high-grade and low-grade gliomas
subjects is the same in each fold. 
The scans are labelled with five classes (Fig.~\ref{fig:tree}): (0)
background, (1) necrotic core, (2) edema, (3) non-enhancing core and
(4) enhancing tumour. The most common evaluation criteria for BraTS is
to use the Dice scores for the whole tumour (labels 1,2,3,4), the core
tumour (labels 1,3,4) and the enhanced tumour (label 4). 
All the scans of BraTS dataset are skull stripped,
resampled to a 1mm isotropic grid and co-registered to the T1-weighted volume of each patient. 
Additionally, we applied histogram standardisation to each
imaging modality independently~\cite{histStd}.

\subsection{Implementation details}
We train the networks using ADAM~\cite{adam} with a
learning rate $lr=0.01$, $\beta_1 = 0.9$ and $\beta_2 = 0.999$. To
regularise the network, we use early stopping on the validation set
and dropout in all residual blocks before the last activation (as
proposed in~\cite{wideresnet}), with a probability of $0.6$. 
We use multi-modal volumes of size $80^3$ from one
subject concatenated as input during training and a
sampling strategy to maximise the number of
classes in each patch. 
Experiments have been performed using Tensorflow
1.1~\footnote{The code is publicly available as part of
  NiftyNet (http://niftynet.io)} and a Nvidia GeForce GTX Titan X GPU.

\section{Results}

We evaluate the usefulness of the proposed soft multi-class Wasserstein Dice loss and the proposed HCNN with deep supervision. We compare the soft multi-class Wasserstein Dice loss to the state-of-the-art mean Dice score~\cite{scalenet,highresnet} for the training of our HCNN in Table \ref{tab:loss_results} and \ref{tab:confusion}. We also evaluate the segmentation at the different scales of the HCNN in Table \ref{tab:holistic_results}.

%

\begin{table}[tb]
	\centering
	\caption{Evaluation of different multi-class Dice scores for training and testing. $\mathcal{L}_{D^{M_{tree}}-PT}$ stands for pre-training the HCNN with mean Dice score (4 epochs) and retraining it with $\mathcal{L}_{D^{M_{tree}}}$ (85 epochs).}
	\begin{tabularx}{\textwidth}{c *{6}{Y}}
		\toprule
		\multicolumn{1}{c}{\bf Loss function}
		& \multicolumn{6}{c}{\bf Evaluation: Mean(std) Dice scores (\%)}\\
		\cmidrule(lr){2-7}
		 & Whole & Core & Enh. & Mean Dice & $D^{M_{0-1}}$  & $D^{M_{tree}}$\\ 
		\midrule
		Mean Dice                  & 83(13) & 70(21) & 68(26) & \bf60(12) & 77(11) & 80(12)\\
		$\mathcal{L}_{D^{M_{0-1}}}$ & 86(12) & 59(29) & 69(23) & 48(5) & 82(6) & 85(5)\\
		$\mathcal{L}_{D^{M_{tree}}}$ & 88(8) & 73(23) & 70(25) & 54(7) & 84(5) & 86(5)\\
		$\mathcal{L}_{D^{M_{tree}}-PT}$ & \bf89(6) & \bf73(22) & \bf74(23) & 59(10) & \bf84(4) & \bf87(4)\\
		\bottomrule
	\end{tabularx}
	\label{tab:loss_results}
\end{table}

\subsection{Examples of distance metrics on BraTS label space}
To illustrate the flexibility of the proposed generalised Wasserstein
Dice score, we evaluate two semantically driven choices for the
distance matrix $M$ on $\mathbf{L}$: 
\[
M_{0-1}= \begin{pmatrix}
0  & 1 & 1 & 1 & 1 \\
1  & 0 & 1 & 1 & 1 \\
1  & 1 & 0 & 1 & 1 \\
1  & 1 & 1 & 0 & 1 \\
1  & 1 & 1 & 1 & 0 \\
\end{pmatrix}, \quad \textrm{and} \quad
M_{tree}= \begin{pmatrix}
0  & 1 & 1 & 1 & 1 \\
1  & 0 & 0.6 & 0.2 & 0.5 \\
1  & 0.6 & 0 & 0.6 & 0.7 \\
1  & 0.2 & 0.6 & 0 & 0.5 \\
1  & 0.5 & 0.7 & 0.5 & 0 \\
\end{pmatrix}.
\]
$M_{0-1}$ is associated with the discrete distance on $\mathbf{L}$ with no inter-class relationship. 
%
%
$M_{tree}$ is derived from the tree structure of $\mathbf{L}$
illustrated in Fig.~\ref{fig:tree}.
This tree is based on the tumour hierarchical structure: whole, core and enhancing tumour. We set branch weights to $0.1$ for contiguous nodes and $0.2$ otherwise.

\subsection{Evaluation and training with multi-class Dice score}

\begin{figure}[t!]
	\centering
	\includegraphics[width=\linewidth]{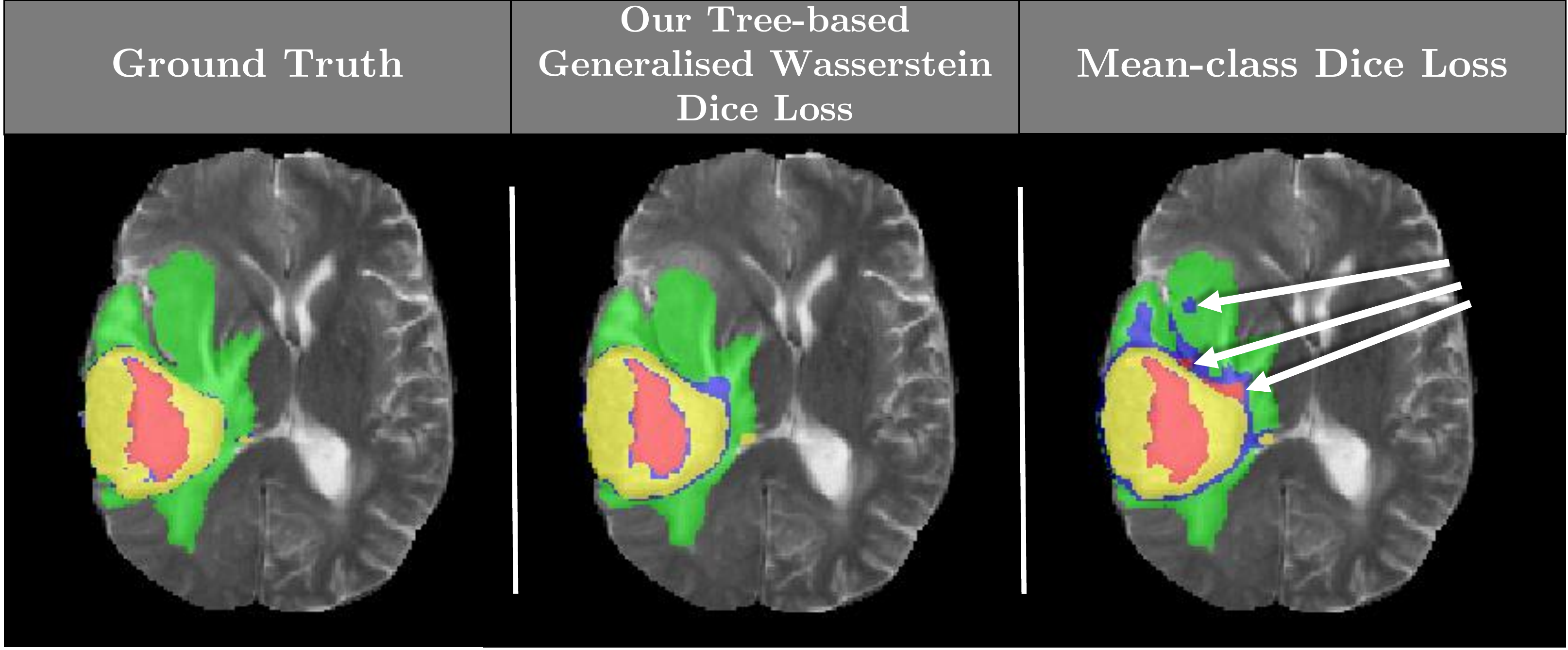}
	\caption{Qualitative comparison of HCNN predictions at testing after training with the proposed Generalised Wasserstein Dice loss ($\mathcal{L}_{D^{M_{tree}}-PT}$) or mean-class Dice loss. Training with $\mathcal{L}_{D^{M_{tree}}-PT}$ allows avoiding implausible misclassifications encountered in predictions after training with mean-class Dice loss (emphasized by white arrows).}
	\label{fig:loss_quali}
\end{figure}
The mean Dice corresponds to the mean of soft Dice scores for each class as used in~\cite{scalenet,highresnet}.
Results in Table~\ref{tab:loss_results} confirm that training with mean Dice score, $D^{M_{0-1}}$ or 
 $D^{M_{tree}}$ allow maximising results for the associated multi-class Dice score during inference.

While $D^{M_{tree}}$ takes advantage of prior information about the hierarchical structure of the tumour classes it makes the optimisation more complex by adding more constraints.
To relax those constraints, we propose to pretrain the network using the mean Dice score during a few epochs (4 in our experiment) and then retrain it using $D^{M_{tree}}$.
This approach leads to the best results for all criteria, as illustrated in the last line of Table~\ref{tab:loss_results}.
Moreover, it produces segmentations that are more semantically plausible compared to the HCNN trained with mean Dice only as illustrated by Fig.~\ref{fig:loss_quali}.

\subsection{Impact of the Wasserstein Dice loss on class confusion}

\begin{table}[t]
	\centering
	\caption{Dice score evaluation of the confusion after training the HCNN using different loss functions. Each line (resp. column) corresponds to the mean(standard deviation) Dice scores (\%) of a region of the ground truth (resp. prediction) with all regions of the prediction (resp. ground truth) computed on the testing set.}
	\begin{tabular}{l | C{1.9cm} | C{1.9cm} | C{1.8cm} | C{1.8cm} | C{1.8cm}}
		\hline
		Mean Dice & \multicolumn{5}{c}{\bf Prediction}\\
		\hline
		\bf Ground truth & Background & Necrotic core & Edema & Non-enh. & Enh. \\ 
		\hline
		Background     & 99.6(0) & 0(0)     &  0.8(0)   & 0.1(0)  & 0.1(0)     \\
		Necrotic core  & 0.(0)   & 36.8(30) &  1.2(3)   & 8.3(9)  & 1(1)   \\
		Edema          & 0.3(0)  & 0.9(1)   &  62.9(18) & 21.7(13)& 4.3(6)   \\
		Non-enh.       & 0.1(0)  & 8.7(9)   &  6.5(8)   & 33(15)  & 14.8(11)   \\
		Enh.           & 0(0)    & 0.9(1)   &  0.3(0)   & 6.9(7)  & 67.6(25) \\
		\hline	
	\end{tabular}
	\vspace{0.3cm}
	
	\begin{tabular}{l| C{1.9cm} | C{1.9cm} | C{1.8cm} | C{1.8cm} | C{1.8cm}}
		\hline
		$\mathcal{L}_{D^{M_{tree}}}$  & \multicolumn{5}{c}{\bf Prediction}\\
		\hline
		\bf Ground truth & Background & Necrotic core & Edema & Non-enh. & Enh. \\ 
		\hline
		Background     & 99.7(0) & 0(0) &  0.3(0)   & 0(0)     & 0(0)     \\
		Necrotic core  & 0(0)    & 0(0) &  2.4(5)   & 28.2(22) & 1.4(1)   \\
		Edema          & 0.6(0)  & 0(0) &  71.3(12) & 8.5(7)   & 3.5(5)   \\
		Non-enh.       & 0.1(0)  & 0(0) &  15.4(13) & 28.9(14) & 14.2(10) \\
		Enh.           & 0(0)    & 0(0) &  1.7(1)   & 6.9(7)   & 70.5(25) \\
		\hline
	\end{tabular}
	\vspace{0.3cm}
	
	\begin{tabular}{l| C{1.9cm} | C{1.9cm} | C{1.8cm} | C{1.8cm} | C{1.8cm}}
		\hline
		$\mathcal{L}_{D^{M_{tree}}-PT}$  & \multicolumn{5}{c}{\bf Prediction}\\
		\hline
		\bf Ground truth & Background & Necrotic core & Edema & Non-enh. & Enh. \\ 
		\hline
		Background     & 99.7(0) & 0(0)     &  0.2(0)   & 0(0)     & 0(0)     \\
		Necrotic core  & 0(0)    & 20.2(27) &  2.3(5)   & 23.2(18) & 1(1)   \\
		Edema          & 0.6(0)  & 0.3(0)   &  73.3(11) & 5.7(5)   & 3.1(4)   \\
		Non-enh.       & 0.1(0)  & 2.1(7)   &  16.1(13) & 30(17)   & 13(8) \\
		Enh.           & 0(0)    & 0(0)     &  2.4(2)   & 3.8(4)   & 73.5(22) \\
		\hline
	\end{tabular}
	\label{tab:confusion}
\end{table}

Evaluating brain tumour segmentation using Dice scores of label subsets like whole, core and enhancing tumour doesn't allow measuring the ability of a model to learn inter-class relationships and to favour voxel classifications, be it correct or not, that are semantically as close as possible to the ground truth.
We propose to measure class confusion using pairwise comparisons of all labels pair between the predicted segmentation and the ground truth (Table~\ref{tab:confusion}). Mathematically, for all $l,l' \in \mathbf{L}$, the quantity in row $l$ and colomn $l'$ stands for the soft binary Dice score:
\begin{equation}
	D_{l,l'} = \frac{2\sum_i g^i_lp^i_{l'}}{\sum_i (g^i_l + p^i_{l'})}
\end{equation}
Results in Table~\ref{tab:confusion} compare class confusion of the proposed HCNN after being trained either using mean Dice loss, tree-based Wasserstein Dice loss ($\mathcal{L}_{D^{M_{tree}}}$) or tree-based Wasserstein Dice loss pre-trained with mean Dice loss($\mathcal{L}_{D^{M_{tree}}-PT}$). The first one aims only at maximising the true positives (diagonal) while the two other additionally aim at balancing the misclassifications to produce semnatically meaningful segmentations.

The network trained with mean Dice loss segments correctly most of the voxels (diagonal in Table~\ref{tab:confusion}) but makes misclassifications that are not semantically meaningful. For example, it makes poor differentiation between the edema and the core tumour as can be seen in the line corresponding to edema in Table~\ref{tab:confusion} and in Fig.~\ref{fig:loss_quali}. 

In contrast, the network trained with $\mathcal{L}_{D^{M_{tree}}}$ makes more meaningful confusion but it is not able to differentiate necrotic core and non-enhancing tumour at all (columns 2 and 4). It illustrates the difficulty to train the network with $\mathcal{L}_{D^{M_{tree}}}$ starting from a random initialisation because $\mathcal{L}_{D^{M_{tree}}}$ embeds more constraints than the mean Dice loss.

$\mathcal{L}_{D^{M_{tree}}-PT}$ allows combining advantages of both loss function:
pre-training the network using the mean Dice loss allows initialising it so that it produces quickly an approximation of the segmentation, and retraining it with $\mathcal{L}_{D^{M_{tree}}}$ allows reaching a model which provides semantically meaningful segmentations (Fig.~\ref{fig:loss_quali}) with a higher rate of true positives compared to training with $\mathcal{L}_{D^{M_{tree}}}$ or mean Dice loss alone (Table~\ref{tab:confusion}).

\subsection{Evaluation of deep supervision}

\begin{table}[tb]
	\centering
	\caption{Evaluation of scale-specific and fused predictions of the HCNN with Dice score of whole, core, enhancing tumour and $D^{M_{tree}}$ after being pre-trained with mean Dice score (4 epochs) and retrained with $\mathcal{L}_{D^{M_{tree}}}$ (85 epochs).}
	\begin{tabularx}{0.9\textwidth}{c *{5}{Y}}
		\toprule
		 \bf Prediction & \multicolumn{4}{c}{\bf Mean(Std) Dice score (\%)}\\
		\cmidrule(lr){2-5}
		 & Whole tumour & Core tumour & Enh. tumour & $D^{M_{tree}}$\\ 
		\midrule
		Scale 1 & 84(8)    & 68(23)   & 70(25)    & 84(5)\\
		Scale 2 & \bf89(5) & \bf73(22) & \bf74(23)  & \bf87(4)\\
		Scale 3 & 88(6)    & 72(23)   & 71(22)    & 86(4)\\
		Scale 4 & \bf89(5) & 72(22) & 71(21) & 86(3)\\
		Fused   & 89(6) & \bf73(22) & \bf74(23) & \bf87(4)\\
		\bottomrule
	\end{tabularx}
	\label{tab:holistic_results}
\end{table}

\begin{figure}[tb]
	\centering
	\includegraphics[width=\linewidth]{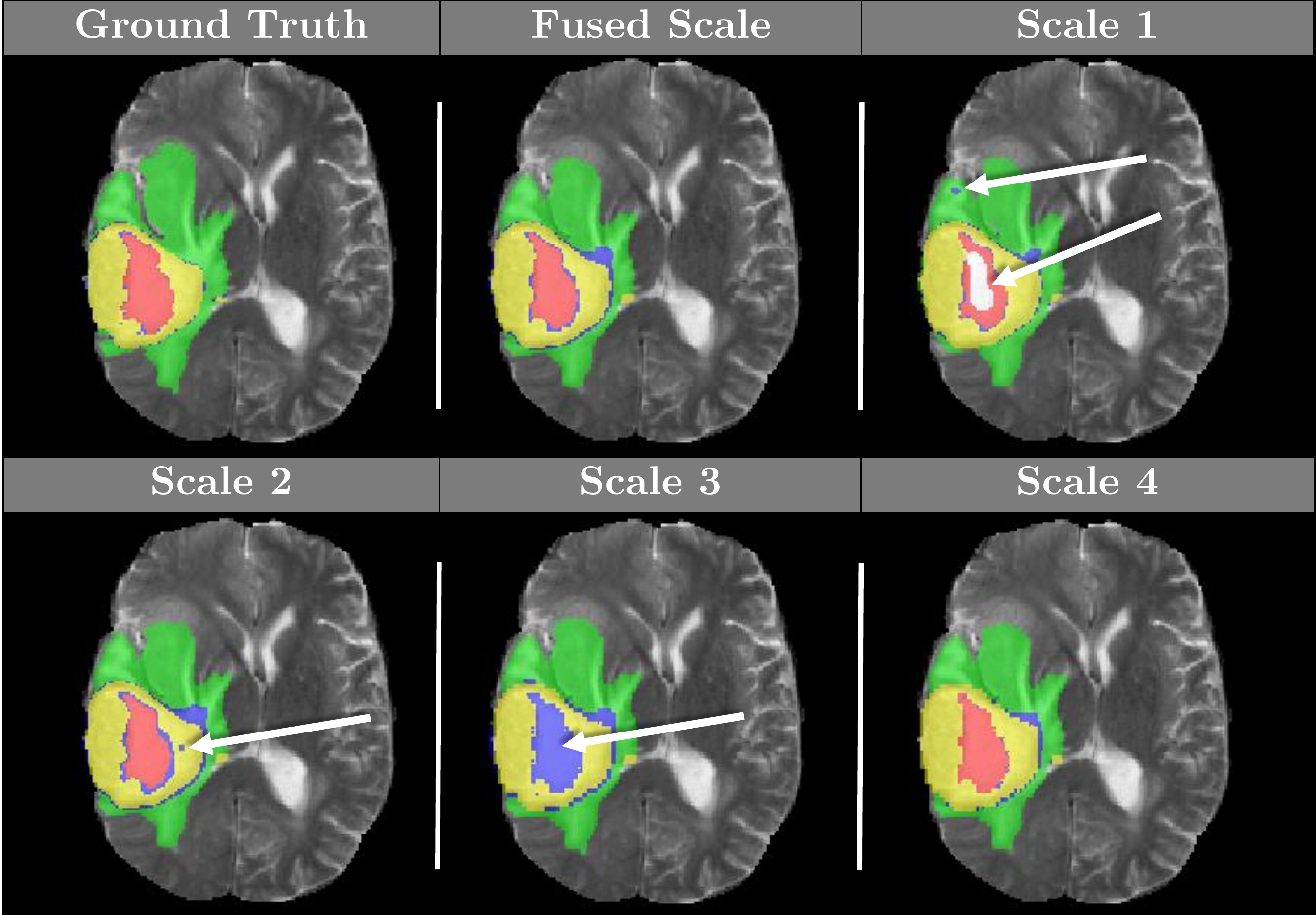}
	\caption{Qualitative comparison of fused and scales predictions at testing after training our HCNN with the proposed Generalised Wasserstein Dice loss ($\mathcal{L}_{D^{M_{tree}}-PT}$). White arrows emphasize implausible misclassifications.}
	\label{fig:scales_quali}
\end{figure}

Results in Table~\ref{tab:holistic_results} are obtained after pre-training HCNN with mean Dice score during 4 epochs and then training it with $\mathcal{L}_{D^{M_{tree}}}$ during 85 additional epochs.
Scales 2 to 4 and fused achieve similar Dice scores for whole, core tumour and the objective function $D^{M_{tree}}$ while scale 1 obtains lower Dice scores. Holes in tumour segmentations produced by scale 1, as illustrated in Fig.~\ref{fig:scales_quali}, suggest an unsufficient receptive field could account for those lower Dice scores.
The best result for the enhancing tumour is achievied by both scale 2 and fused, which was expected as this is the smallest region of interest and the full resolution is maintained until scale 2.
Moreover, as illustrated in Fig.~\ref{fig:scales_quali}, scales 3 and 4 fail at segmenting the thinest regions of the tumour because of their lower resolution contrary to scales 1 and 2 and fused.
However, scales 1 to 3 contained implausible segmentation regions contrary to scale 4 and fused.
This suggests trade-offs between high receptive field and high resolution that are class specific. It confirms the usefulness of the multi-scale holistic approach for the multi-class brain tumour segmentation task.


\section{Conclusion and future work}

We proposed a semantically driven generalisation of the Dice score for
soft multi-class segmentation based on the Wasserstein distance. This
embeds prior knowledge about inter-class relationships represented by
a distance  matrix on the label space. 
Additionally, we proposed a holistic convolutional network that uses
multi-scale predictions and deep supervision to make use of multi-scale
information. 
We successfully used the proposed Wasserstein Dice score as a loss
function to train our holistic networks and show the
importance of multi-scale and inter-class relationships for the
imbalanced task of multi-class brain tumour segmentation.
The proposed distance matrix based on the label space tree structure leads to higher Dice scores compared to the discrete distance.
Because the tree-based distance matrix used was heuristically chosen we think that better heuristics or a method to directly learn the matrix from the data could lead to further improvements.

As the memory capacity of GPUs increases, entire multi-modal volumes
could be used as input of CNN-based segmentation. However, it will also increase the class imbalance in the patches used as input.
We expect this to increase the impact of our contributions. 
Future work includes extending the use of Wasserstein distance by
defining a matrix distance on the entire output space
$\mathbf{X}\times\mathbf{L}$ similarly to~\cite{manifold_ot}. This would
allow embedding spatial information directly in the loss, but the
computation burden of the Wasserstein distance, in that case, remains a
challenge~\cite{fast_emd}. 

\subsubsection*{Acknowledgements.}\hspace*{\fill} \\
This work was supported by the Wellcome Trust (WT101957, 203145Z/16/Z, HICF-T4-275, WT 97914), EPSRC (NS/A000027/1, EP/H046410/1, EP/J020990/1, EP/K005278, NS/A000050/1), the NIHR BRC UCLH/UCL, a UCL ORS/GRS Scholarship and a hardware donation from NVidia.


\bibliographystyle{splncs03}
\bibliography{JNAbrv,bibliography}

\begin{thebibliography}{10}
\providecommand{\url}[1]{\texttt{#1}}
\providecommand{\urlprefix}{URL }

\bibitem{Anbeek2005}
Anbeek, P., Vincken, K.L., van Bochove, G.S., van Osch, M.J., van~der Grond,
  J.: Probabilistic segmentation of brain tissue in {MR} imaging. NeuroImage
  27(4),  795 -- 804 (2005)

\bibitem{chang2009performance}
Chang, H.H., Zhuang, A.H., Valentino, D.J., Chu, W.C.: Performance measure
  characterization for evaluating neuroimage segmentation algorithms.
  Neuroimage  (2009)

\bibitem{elu}
Clevert, D.A., Unterthiner, T., Hochreiter, S.: Fast and accurate deep network
  learning by exponential linear units (elus). arXiv:1511.07289  (2015)

\bibitem{gdsc}
Crum, W.R., Camara, O., Hill, D.L.: Generalized overlap measures for evaluation
  and validation in medical image analysis. IEEE TMI  25(11),  1451--1461
  (2006)

\bibitem{scalenet}
Fidon, L., Li, W., Garcia-Peraza-Herrera, L.C., Ekanayake, J., Kitchen, N.,
  Ourselin, S., Vercauteren, T.: Scalable multimodal convolutional networks for
  brain tumour segmentation. MICCAI  (2017)

\bibitem{manifold_ot}
Fitschen, J.H., Laus, F., Schmitzer, B.: Optimal transport for manifold-valued
  images. In: International Conference on Scale Space and Variational Methods
  in Computer Vision. pp. 460--472. Springer (2017)

\bibitem{wasserstein_loss}
Frogner, C., Zhang, C., Mobahi, H., Araya, M., Poggio, T.A.: Learning with a
  wasserstein loss. In: NIPS. pp. 2053--2061 (2015)

\bibitem{toolnet}
Garcia-Peraza-Herrera, L.C., Li, W., Fidon, L., Gruijthuijsen, C., Devreker,
  A., Attilakos, G., Deprest, J., {Vander Poorten}, E., Stoyanov, D.,
  Vercauteren, T., Ourselin, S.: {ToolNet : Holistically-Nested Real-Time
  Segmentation of Robotic Surgical Tools}. In: IROS (2017)

\bibitem{Havaei2017}
Havaei, M., Davy, A., Warde-Farley, D., Biard, A., Courville, A., Bengio, Y.,
  Pal, C., Jodoin, P.M., Larochelle, H.: Brain tumor segmentation with deep
  neural networks. Med. Image Anal.  35,  18--31 (2017)

\bibitem{resnet}
He, K., Zhang, X., Ren, S., Sun, J.: Deep residual learning for image
  recognition. In: IEEE {CVPR} (2016)

\bibitem{deep_medic}
Kamnitsas, K., Ledig, C., Newcombe, V.F., Simpson, J.P., Kane, A.D., Menon,
  D.K., Rueckert, D., Glocker, B.: Efficient multi-scale {3D CNN} with fully
  connected {CRF} for accurate brain lesion segmentation. Med. Image Anal.  36,
   61--78 (2017)

\bibitem{adam}
Kingma, D., Ba, J.: Adam: A method for stochastic optimization. arXiv:1412.6980
   (2014)

\bibitem{highresnet}
Li, W., Wang, G., Fidon, L., Ourselin, S., Cardoso, M.J., Vercauteren, T.: On
  the compactness, efficiency, and representation of {3D} convolutional
  networks: Brain parcellation as a pretext task. IPMI  (2017)

\bibitem{brats}
Menze, B.H., Jakab, A., Bauer, S., Kalpathy-Cramer, J., Farahani, K., Kirby,
  J., Burren, Y., Porz, N., Slotboom, J., Wiest, R., et~al.: The multimodal
  brain tumor image segmentation benchmark ({BraTS}). IEEE TMI  34(10),
  1993--2024 (2015)

\bibitem{v_net}
Milletari, F., Navab, N., Ahmadi, S.A.: {V-Net}: Fully convolutional neural
  networks for volumetric medical image segmentation. In: Proc. {3DV}'16. pp.
  565--571 (2016)

\bibitem{histStd}
Nyul, L.G., Udupa, J.K., Zhang, X.: New variants of a method of {MRI} scale
  standardization. IEEE TMI  19(2),  143--150 (Feb 2000)

\bibitem{fast_emd}
Pele, O., Werman, M.: Fast and robust earth mover's distances. ICCV  (2009)

\bibitem{Sudre2017}
Sudre, C.H., Li, W., Vercauteren, T., Ourselin, S., Cardoso, M.J.: {Generalised
  Dice overlap as a deep learning loss function for highly unbalanced
  segmentations}. arXiv:1707.03237  (2017)

\bibitem{hed}
Xie, S., Tu, Z.: Holistically-nested edge detection. In: ICCV (2015)

\bibitem{wideresnet}
Zagoruyko, S., Komodakis, N.: Wide residual networks. arXiv:1605.07146  (2016)

\end{thebibliography}

\end{document}